# Similarity-based Learning via Data Driven Embeddings

**Purushottam Kar**
Indian Institute of Technology
Kanpur, INDIA
purushot@cse.iitk.ac.in

Prateek Jain
Microsoft Research India
Bangalore, INDIA
prajain@microsoft.com

## Abstract

We consider the problem of classification using similarity/distance functions over data. Specifically, we propose a framework for defining the *goodness* of a (dis)similarity function with respect to a given learning task and propose algorithms that have guaranteed generalization properties when working with such *good* functions. Our framework unifies and generalizes the frameworks proposed by [1] and [2]. An attractive feature of our framework is its adaptability to data - we do not promote a fixed notion of goodness but rather let data dictate it. We show, by giving theoretical guarantees that the goodness criterion best suited to a problem can itself be learned which makes our approach applicable to a variety of domains and problems. We propose a *landmarking*-based approach to obtaining a classifier from such learned goodness criteria. We then provide a novel diversity based heuristic to perform task-driven selection of landmark points instead of random selection. We demonstrate the effectiveness of our goodness criteria learning method as well as the landmark selection heuristic on a variety of similarity-based learning datasets and benchmark UCI datasets on which our method consistently outperforms existing approaches by a significant margin.

## 1  Introduction

Machine learning algorithms have found applications in diverse domains such as computer vision, bio-informatics and speech recognition. Working in such heterogeneous domains often involves handling data that is not presented as explicit features embedded into vector spaces. However in many domains, for example co-authorship graphs, it is natural to devise similarity/distance functions over pairs of points. While classical techniques like decision tree and linear perceptron cannot handle such data, several modern machine learning algorithms such as support vector machine (SVM) can be *kernelized* and are thereby capable of using kernels or similarity functions.

However, most of these algorithms require the similarity functions to be positive semi-definite (PSD), which essentially implies that the similarity stems from an (implicit) embedding of the data into a Hilbert space. Unfortunately in many domains, the most natural notion of similarity does not satisfy this condition - moreover, verifying this condition is usually a non-trivial exercise. Take for example the case of images on which the most natural notions of distance (Euclidean, Earth-mover) [3] do not form PSD kernels. Co-authorship graphs give another such example.

Consequently, there have been efforts to develop algorithms that do not make assumptions about the PSD-ness of the similarity functions used. One can discern three main approaches in this area. The first approach tries to coerce a given similarity measure into a PSD one by either clipping or shifting the spectrum of the kernel matrix [4, 5]. However, these approaches are mostly restricted to transductive settings and are not applicable to large scale problems due to eigenvector computation requirements. The second approach consists of algorithms that either adapt classical methods like $k$-NN to handle non-PSD similarity/distance functions and consequently offer slow test times [5], or are forced to solve non-convex formulations [6, 7].

The third approach, which has been investigated recently in a series of papers [1, 2, 8, 9], uses the similarity function to embed the domain into a low dimensional Euclidean space. More specifically, these algorithms choose *landmark* points in the domain which then give the embedding. Assuming a certain "goodness" property (that is formally defined) for the similarity function, these models offer both generalization guarantees in terms of how well-suited the similarity function is to the classification task as well as the ability to use fast algorithmic techniques such as linear SVM [10] on the *landmarked space*. The model proposed by Balcan-Blum in [1] gives sufficient conditions for a similarity function to be well suited to such a landmarking approach. Wang et al. in [2] on the other hand provide goodness conditions for dissimilarity functions that enable landmarking algorithms.

Informally, a similarity (or distance) function can be said to be good if points of similar labels are closer to each other than points of different labels in some sense. Both the models described above restrict themselves to a fixed goodness criterion, which need not hold for the underlying data. We observe that this might be too restrictive in many situations and present a framework that allows us to tune the goodness criterion itself to the classification problem at hand. Our framework consequently unifies and generalizes those presented in [1] and [2]. We first prove generalization bounds corresponding to landmarked embeddings under a *fixed* goodness criterion. We then provide a uniform-convergence bound that enables us to learn the best goodness criterion for a given problem. We further generalize our framework by giving the ability to incorporate any Lipschitz loss function into our goodness criterion which allows us to give guarantees for the use of various algorithms such as C-SVM and logistic regression on the landmarked space.

Now similar to [1, 2], our framework requires random sampling of training points to create the embedding space[1]. However in practice, random sampling is inefficient and requires sampling of a large number of points to form a useful embedding, thereby increasing training and test time. To address this issue, [2] proposes a heuristic to select the points that are to be used as landmarks. However their scheme is tied to their optimization algorithm and is computationally inefficient for large scale data. In contrast, we propose a general heuristic for selecting informative landmarks based on a novel notion of diversity which can then be applied to any instantiation of our model.

Finally, we apply our methods to a variety of benchmark datasets for similarity learning as well as ones from the UCI repository. We empirically demonstrate that our learning model and landmark selection heuristic consistently offers significant improvements over the existing approaches. In particular, for small number of landmark points, which is a practically important scenario as it is expensive to compute similarity function values at test time, our method provides, on an average, accuracy boosts of upto 5% over existing methods. We also note that our methods can be applied on top of any strategy used to learn the similarity measure (eg. MKL techniques [11]) or the distance measure (eg. [12]) itself. Akin to [1], our techniques can also be extended to learn a combination of (dis)similarity functions but we do not explore these extensions in this paper.

## 2  Methodology

Let $\mathcal{D}$ be a fixed but unknown distribution over the labeled input domain $\mathcal{X}$ and let $\ell : \mathcal{X} \to \{-1, +1\}$ be a labeling over the domain. Given a (potentially non-PSD) similarity function[2] $K : \mathcal{X} \times \mathcal{X} \to \mathbb{R}$, the goal is to learn a classifier $\hat{\ell} : \mathcal{X} \to \{-1, +1\}$ from a finite number of i.i.d. samples from $\mathcal{D}$ that has bounded generalization error over $\mathcal{D}$.

Now, learning a reasonable classifier seems unlikely if the given similarity function does not have any inherent "goodness" property. Intuitively, the goodness of a similarity function should be its suitability to the classification task at hand. For PSD kernels, the notion of goodness is defined in terms of the margin offered in the RKHS [13]. However, a more basic requirement is that the similarity function should preserve affinities among similarly labeled points - that is to say, a good similarity function should not, on an average, assign higher similarity values to dissimilarly labeled points than to similarly labeled points. This intuitive notion of goodness turns out to be rather robust in the sense that all PSD kernels that offer a good margin in their respective RKHSs satisfy some form of this goodness criterion as well [14].

---

[1]Throughout the paper, we use the terms *embedding space* and *landmarked space* interchangeably.

[2]Results described in this section hold for distance functions as well; we present results with respect to similarity functions for sake of simplicity.

Recently there has been some interest in studying different realizations of this general notion of goodness and developing corresponding algorithms that allow for efficient learning with similarity/distance functions. Balcan-Blum in [1] present a goodness criteria in which a good similarity function is considered to be one that, for most points, assigns a greater average similarity to similarly labeled points than to dissimilarly labeled points. More specifically, a similarity function is $(\epsilon, \gamma)$-good if there exists a weighing function $w : \mathcal{X} \to \mathbb{R}$ such that, at least a $(1 - \epsilon)$ probability mass of examples $x \sim \mathcal{D}$ satisfies:

$$\mathbb{E}_{x' \sim \mathcal{D}}\left[w\left(x'\right) K(x, x') | \ell(x') = \ell(x)\right] \geq \mathbb{E}_{x' \sim \mathcal{D}}\left[w\left(x'\right) K(x, x') | \ell(x') \neq \ell(x)\right] + \gamma. \tag{1}$$

where instead of average similarity, one considers an average weighted similarity to allow the definition to be more general.

Wang *et al* in [2] define a distance function $d$ to be good if a large fraction of the domain is, on an average, closer to similarly labeled points than to dissimilarly labeled points. They allow these averages to be calculated based on some distribution distinct from $\mathcal{D}$, one that may be more suited to the learning problem. However it turns out that their definition is equivalent to one in which one again assigns weights to domain elements, as done by [1], and the following holds

$$\mathbb{E}_{x', x'' \sim \mathcal{D} \times \mathcal{D}}\left[w(x')w(x'') \operatorname{sgn}\left(d(x, x'') - d(x, x')\right) | \ell(x') = \ell(x), \ell(x'') \neq \ell(x)\right] > \gamma \tag{2}$$

Assuming their respective goodness criteria, [1] and [2] provide efficient algorithms to learn classifiers with bounded generalization error. However these notions of goodness with a single fixed criterion may be too restrictive in the sense that the data and the (dis)similarity function may not satisfy the underlying criterion. This is, for example, likely in situations with high intra-class variance. Thus there is need to make the goodness criterion more flexible and data-dependent.

To this end, we unify and generalize both the above criteria to give a notion of goodness that is more data dependent. Although the above goodness criteria (1) and (2) seem disparate at first, they can be shown to be special cases of a generalized framework where an antisymmetric function is used to compare intra and inter-class affinities. We use this observation to define our novel goodness criterion using arbitrary bounded antisymmetric functions which we refer to as *transfer functions*. This allows us to define a family of goodness criteria of which (1) and (2) form special cases ((1) uses the identity function and (2) uses the sign function as transfer function). Moreover, the resulting definition of a good similarity function is more flexible and data dependent. In the rest of the paper we shall always assume that our similarity functions are normalized i.e. for the domain of interest $\mathcal{X}$, $\sup_{x,y \in \mathcal{X}} K(x, y) \leq 1$.

**Definition 1** (Good Similarity Function). *A similarity function $K : \mathcal{X} \times \mathcal{X} \to \mathbb{R}$ is said to be an $(\epsilon, \gamma, B)$-good similarity for a learning problem where $\epsilon, \gamma, B > 0$ if for some antisymmetric transfer function $f : \mathbb{R} \to \mathbb{R}$ and some weighing function $w : \mathcal{X} \times \mathcal{X} \to [-B, B]$, at least a $(1 - \epsilon)$ probability mass of examples $x \sim \mathcal{D}$ satisfies*

$$\mathbb{E}_{x', x'' \sim \mathcal{D} \times \mathcal{D}}\left[w\left(x', x''\right) f\left(K(x, x') - K(x, x'')\right) | \ell(x') = \ell(x), \ell(x'') \neq \ell(x)\right] \geq C_f \gamma \tag{3}$$

*where* $C_f = \sup_{x, x' \in \mathcal{X}} f(K(x, x')) - \inf_{x, x' \in \mathcal{X}} f(K(x, x'))$

As mentioned before, the above goodness criterion generalizes the previous notions of goodness [3] and is adaptive to changes in data as it allows us, as shall be shown later, to learn the best possible criterion for a given classification task by choosing the most appropriate transfer function from a parameterized family of functions. We stress that the property of antisymmetry for the transfer function is crucial to the definition in order to provide a uniform treatment to points of all classes as will be evident in the proof of Theorem 2.

As in [1, 2], our goodness criterion lends itself to a simple learning algorithm which consists of choosing a set of $d$ random pairs of points from the domain $\mathcal{P} = \left\{\left(x_i^+, x_i^-\right)\right\}_{i=1}^d$ (which we refer to as *landmark pairs*) and defining an embedding of the domain into a *landmarked space* using these landmarks : $\Phi_L : \mathcal{X} \to \mathbb{R}^d, \Phi_L(x) = \left(f(K(x, x_i^+) - K(x, x_i^-))\right)_{i=1}^d \in \mathbb{R}^d$. The advantage of performing this embedding is the guaranteed existence of a large margin classifier in the landmarked space as shown below.

---

[3]We refer the reader to the appendix for a discussion.

**Theorem 2.** *If $K$ is an $(\epsilon, \gamma, B)$-good similarity with respect to transfer function $f$ and weight function $w$ then for any $\epsilon_1 > 0$, with probability at least $1 - \delta$ over the choice of $d = (8/\gamma^2) \ln(2/\delta\epsilon_1)$ positive and negative samples, $\{x_i^+\}_{i=1}^d \subset \mathcal{D}^+$ and $\{x_i^-\}_{i=1}^d \subset \mathcal{D}^-$ respectively, the classifier $h(x) = sgn[g(x)]$ where $g(x) = \frac{1}{d} \sum_{i=1}^d w(x_i^+, x_i^-) f\left(K(x, x_i^+) - K(x, x_i^-)\right)$ has error no more than $\epsilon + \epsilon_1$ at margin $\frac{\gamma}{2}$.*

*Proof.* We shall prove that with probability at least $1 - \delta$, at least a $1 - \epsilon_1$ fraction of points $x$ that satisfy Equation 3 are classified correctly by the classifier $h(x)$. Overestimating the error by treating the points that do not satisfy Equation 3 as always being misclassified will give us the desired result.

For any fixed $x \in \mathcal{X}^+$ that satisfies Equation 3, we have

$$\mathbb{E}_{x', x'' \sim \mathcal{D} \times \mathcal{D}} \left[w(x', x'') f\left(K(x, x') - K(x, x'')\right) | \ell(x') = 1, \ell(x'') = -1\right] \geq C_f \gamma$$

hence the Hoeffding Bounds give us

$$\Pr\left[g(x) < \frac{\gamma}{2}\right] = \Pr\left[\frac{1}{d} \sum_{i=1}^d w(x_i^+, x_i^-) f\left(K(x, x_i^+) - K(x, x_i^-)\right) < \frac{\gamma}{2}\right] \leq 2 \exp\left(-\frac{\gamma^2 d}{8}\right)$$

Similarly, for any fixed $x \in \mathcal{X}^-$ that satisfies Equation 3, we have

$$\mathbb{E}_{x', x'' \sim \mathcal{D} \times \mathcal{D}} \left[w(x', x'') f\left(K(x, x') - K(x, x'')\right) | \ell(x') = -1, \ell(x'') = 1\right] \geq C_f \gamma$$

hence the Hoeffding Bounds give us

$$\begin{aligned}\Pr\left[g(x) > \frac{\gamma}{2}\right] &= \Pr\left[\frac{1}{d} \sum_{i=1}^d w(x_i^+, x_i^-) f\left(K(x, x_i^+) - K(x, x_i^-)\right) > \frac{\gamma}{2}\right] \\ &= \Pr\left[\frac{1}{d} \sum_{i=1}^d w(x_i^+, x_i^-) f\left(K(x, x_i^-) - K(x, x_i^+)\right) < \frac{\gamma}{2}\right] \leq 2 \exp\left(-\frac{\gamma^2 d}{8}\right)\end{aligned}$$

where in the second step we have used antisymmetry of $f$.

Since we have shown that this result holds true individually for any point $x$ that satisfies Equation 3, the expected error (where the expectation is both over the choice of domain points as well as choice of the landmark points) itself turns out to be less than $2 \exp\left(-\frac{\gamma^2 d}{8}\right) \leq \epsilon_1 \delta$. Applying Markov's inequality gives us that the probability of obtaining a set of landmarks such that the error on points satisfying Equation 3 is greater than $\epsilon_1$ is at most $\delta$.

Assuming, as mentioned earlier, that the points not satisfying Equation 3 can always be misclassified proves our desired result. □

However, there are two hurdles to obtaining this large margin classifier. Firstly, the existence of this classifier itself is predicated on the use of the correct transfer function, something which is unknown. Secondly, even if an optimal transfer function is known, the above formulation cannot be converted into an efficient learning algorithm for discovering the (unknown) weights since the formulation seeks to minimize the number of misclassifications which is an intractable problem in general.

We overcome these two hurdles by proposing a nested learning problem. First of all we assume that for some fixed loss function $L$, given any transfer function and any set of landmark pairs, it is possible to obtain a large margin classifier in the corresponding landmarked space that minimizes $L$. Having made this assumption, we address below the issue of learning the optimal transfer function for a given learning task. However as we have noted before, this assumption is not valid for arbitrary loss functions. This is why, subsequently in Section 2.2, we shall show it to be valid for a large class of loss functions by incorporating surrogate loss functions into our goodness criterion.

## 2.1 Learning the transfer function

In this section we present results that allow us to learn a near optimal transfer function from a family of transfer functions. We shall assume, for some fixed loss function $L$, the existence of an efficient routine which we refer to as TRAIN that shall return, for any landmarked space indexed by a set of landmark pairs $\mathcal{P}$, a large margin classifier minimizing $L$. The routine TRAIN is allowed to make use of additional training data to come up with this classifier.

An immediate algorithm for choosing the best transfer function is to simply search the set of possible transfer functions (in an algorithmically efficient manner) and choose the one offering lowest training error. We show here that given enough landmark pairs, this simple technique, which we refer to as FTUNE (see Algorithm 2) is guaranteed to return a near-best transfer function. For this we prove a uniform convergence type guarantee on the space of transfer functions.

Let $\mathcal{F} \subset [-1,1]^{\mathbb{R}}$ be a class of antisymmetric functions and $\mathcal{W} = [-B, B]^{\mathcal{X} \times \mathcal{X}}$ be a class of weight functions. For two real valued functions $f$ and $g$ defined on $\mathcal{X}$, let $\|f - g\|_\infty := \sup_{x \in \mathcal{X}} |f(x) - g(x)|$. Let $\mathcal{B}_\infty(f, r) := \{f' \in \mathcal{F} \mid \|f - f'\|_\infty < r\}$. Let $L$ be a $C_L$-Lipschitz loss function. Let $\mathcal{P} = \{(x_i^+, x_i^-)\}_{i=1}^d$ be a set of (random) landmark pairs. For any $f \in \mathcal{F}, w \in \mathcal{W}$, define

$$G_{(f,w)}(x) = \mathop{\mathbb{E}}_{x',x'' \sim \mathcal{D} \times \mathcal{D}} \left[ w(x', x'') f(K(x, x') - K(x, x'')) \mid \ell(x') = \ell(x), \ell(x'') \neq \ell(x) \right]$$

$$g_{(f,w)}(x) = \frac{1}{d} \sum_{i=1}^{d} w(x_i^+, x_i^-) f(K(x, x_i^+) - K(x, x_i^-))$$

Theorem 7 (see Section 2.2) guarantees us that for any fixed $f$ and any $\epsilon_1 > 0$, if $d$ is large enough then $\mathop{\mathbb{E}}_x [L(g_{(f,w)}(x))] \leq \mathop{\mathbb{E}}_x [L(G_{(f,w)}(x))] + \epsilon_1$. We now show that a similar result holds even if one is allowed to vary $f$. Before stating the result, we develop some notation.

For any transfer function $f$ and arbitrary choice of landmark pairs $\mathcal{P}$, let $w_{(g,f)}$ be the *best* weighing function for this choice of transfer function and landmark pairs i.e. let $w_{(g,f)} = \arg\min_{w \in [-B,B]^d} \mathop{\mathbb{E}}_{x \sim \mathcal{D}} [L(g_{(f,w)}(x))]$ [4]. Similarly, let $w_{(G,f)}$ be the best weighing function corresponding to $G$ i.e. $w_{(G,f)} = \arg\min_{w \in \mathcal{W}} \mathop{\mathbb{E}}_{x \sim \mathcal{D}} [L(G_{(f,w)}(x))]$. Then we can ensure the following:

**Theorem 3.** *Let $\mathcal{F}$ be a compact class of transfer functions with respect to the infinity norm and $\epsilon_1, \delta > 0$. Let $\mathcal{N}(\mathcal{F}, r)$ be the size of the smallest $\epsilon$-net over $\mathcal{F}$ with respect to the infinity norm at scale $r = \frac{\epsilon_1}{4C_L B}$. Then if one chooses $d = \frac{64 B^2 C_L^2}{\epsilon_1^2} \ln\left(\frac{16 B \cdot \mathcal{N}(\mathcal{F}, r)}{\delta \epsilon_1}\right)$ random landmark pairs then we have the following with probability greater than $(1 - \delta)$*

$$\sup_{f \in \mathcal{F}} \left[ \left| \mathop{\mathbb{E}}_{x \sim \mathcal{D}} \left[ L\left( g_{(f, w_{(g,f)})}(x) \right) \right] - \mathop{\mathbb{E}}_{x \sim \mathcal{D}} \left[ L\left( G_{(f, w_{(G,f)})}(x) \right) \right] \right| \right] \leq \epsilon_1$$

We shall prove the theorem in two parts. As we shall see, one of the parts is fairly simple to prove. To prove the other part, we shall exploit the Lipschitz properties of the loss function as well as the fact that the class of transfer functions chosen form a compact set. Let us call a given set of landmark pairs to be *good* with respect to a fixed transfer function $f \in \mathcal{F}$ if for the corresponding $g$, $\mathop{\mathbb{E}}_x [L(g(x))] \leq \mathop{\mathbb{E}}_x [L(G(x))] + \epsilon_1$ for some small fixed $\epsilon_1 > 0$.

We will first prove, using Lipschitz properties of the loss function that if a given set of landmarks is good with respect to a given transfer function, then it is also good with respect to all transfer functions in its neighborhood. Having proved this, we will apply a standard covering number argument in which we will ensure that a large enough set of landmarks is good with respect to a set of transfer functions that form an $\epsilon$-net over $\mathcal{F}$ and use the previous result to complete the proof.

We first prove a series of simple results which will be used in the first part of the proof. In the following $f$ and $f'$ are two transfer functions such that $f' \in \mathcal{B}_\infty(f, r) \cap \mathcal{F}$.

**Lemma 4.** *The following results are true*

---
[4] Note that the function $g_{(f,w)}(x)$ is dictated by the choice of the set of landmark pairs $\mathcal{P}$

1. *For any fixed $f \in F$, $\mathop{\mathbb{E}}_{x \sim \mathcal{D}}\left[L\left(G_{(f,w_{(G,f)})}(x)\right)\right] \leq \mathop{\mathbb{E}}_{x \sim \mathcal{D}}\left[L\left(G_{(f,w)}(x)\right)\right]$ for all $w \in \mathcal{W}$.*

2. *For any fixed $f \in F$, any fixed $g$ obtained by an arbitrary choice of landmark pairs, $\mathop{\mathbb{E}}_{x \sim \mathcal{D}}\left[L\left(g_{(f,w_{(g,f)})}(x)\right)\right] \leq \mathop{\mathbb{E}}_{x \sim \mathcal{D}}\left[L\left(g_{(f,w)}(x)\right)\right]$ for all $w \in \mathcal{W}$.*

3. *For any $f' \in \mathcal{B}_\infty(f,r) \cap \mathcal{F}$, $\left|\mathop{\mathbb{E}}_{x \sim \mathcal{D}}\left[L\left(G_{(f,w_{(G,f)})}(x)\right)\right] - \mathop{\mathbb{E}}_{x \sim \mathcal{D}}\left[L\left(G_{(f',w_{(G,f')})}(x)\right)\right]\right| \leq C_L r B$.*

4. *For any fixed $g$ obtained by an arbitrary choice of landmark pairs, $f' \in \mathcal{B}_\infty(f,r) \cap \mathcal{F}$, $\left|\mathop{\mathbb{E}}_{x \sim \mathcal{D}}\left[L\left(g_{(f,w_{(g,f)})}(x)\right)\right] - \mathop{\mathbb{E}}_{x \sim \mathcal{D}}\left[L\left(g_{(f',w_{(g,f')})}(x)\right)\right]\right| \leq C_L r B$.*

*Proof.* We prove the results in order,

1. Immediate from the definition of $w_{(G,f)}$.

2. Immediate from the definition of $w_{(g,f)}$.

3. We have $\mathop{\mathbb{E}}_{x \sim \mathcal{D}}\left[L\left(G_{(f',w_{(G,f')})}(x)\right)\right] \leq \mathop{\mathbb{E}}_{x \sim \mathcal{D}}\left[L\left(G_{(f',w_{(G,f)})}(x)\right)\right]$ by an application of Lemma 4.1 proven above. For sake of simplicity let us denote $w_{(G,f)} = w$ for the next set of calculations. Now we have

$$\begin{aligned}
G_{(f',w)}(x) &= \mathop{\mathbb{E}}_{x',x'' \sim \mathcal{D} \times \mathcal{D}} \left[w(x',x'') f'\left(K(x,x') - K(x,x'')\right) | \ell(x') = \ell(x), \ell(x'') \neq \ell(x)\right] \\
&\leq \mathop{\mathbb{E}}_{x',x'' \sim \mathcal{D} \times \mathcal{D}} \left[w(x',x'') \left(f\left(K(x,x') - K(x,x'')\right) + r\right) | \ell(x') = \ell(x), \ell(x'') \neq \ell(x)\right] \\
&= \mathop{\mathbb{E}}_{x',x'' \sim \mathcal{D} \times \mathcal{D}} \left[w(x',x'') f\left(K(x,x') - K(x,x'')\right) | \ell(x') = \ell(x), \ell(x'') \neq \ell(x)\right] \\
&\quad + r \cdot \mathop{\mathbb{E}}_{x',x'' \sim \mathcal{D} \times \mathcal{D}} \left[w(x',x'') | \ell(x') = \ell(x), \ell(x'') \neq \ell(x)\right] \\
&\leq G_{(f,w)}(x) + rB
\end{aligned}$$

where in the second inequality we have used the fact that $\|f - f'\|_\infty \leq r$ and in the fourth inequality we have used the fact that $w \in \mathcal{W}$. Thus we have $G_{(f',w)}(x) \leq G_{(f,w)}(x) + rB$. Using the Lipschitz properties of $L$ we can now get $\mathop{\mathbb{E}}_{x \sim \mathcal{D}}\left[L\left(G_{(f',w)}(x)\right)\right] \leq \mathop{\mathbb{E}}_{x \sim \mathcal{D}}\left[L\left(G_{(f,w)}(x)\right)\right] + C_L r B$. Thus we have $\mathop{\mathbb{E}}_{x \sim \mathcal{D}}\left[L\left(G_{(f',w_{(G,f')})}(x)\right)\right] \leq \mathop{\mathbb{E}}_{x \sim \mathcal{D}}\left[L\left(G_{(f',w_{(G,f)})}(x)\right)\right] \leq \mathop{\mathbb{E}}_{x \sim \mathcal{D}}\left[L\left(G_{(f,w_{(G,f)})}(x)\right)\right] + C_L r B$.

Similarly we can also prove $\mathop{\mathbb{E}}_{x \sim \mathcal{D}}\left[L\left(G_{(f,w_{(G,f)})}(x)\right)\right] \leq \mathop{\mathbb{E}}_{x \sim \mathcal{D}}\left[L\left(G_{(f',w_{(G,f')})}(x)\right)\right] + C_L r B$. This gives us the desired result.

4. The proof follows in a manner similar to the one for Lemma 4.3 proven above. □

Using the above results we get a preliminary form of the first part of our proof as follows :

**Lemma 5.** *Suppose a set of landmarks is $(\epsilon_1/2)$-good for a particular landmark $f \in \mathcal{F}$ (i.e. $\mathop{\mathbb{E}}_{x \sim \mathcal{D}}\left[L\left(g_{(f,w_{(G,f)})}(x)\right)\right] < \mathop{\mathbb{E}}_{x \sim \mathcal{D}}\left[L\left(G_{(f,w_{(G,f)})}(x)\right)\right] + \epsilon_1/2$), then the same set of landmarks are also $\epsilon_1$-good for any $f' \in \mathcal{B}_\infty(f,r) \cap \mathcal{F}$ (i.e. for all $f' \in \mathcal{B}_\infty(f,r) \cap \mathcal{F}$, $\mathop{\mathbb{E}}_{x \sim \mathcal{D}}\left[L\left(g_{(f',w_{(g,f')})}(x)\right)\right] \leq \mathop{\mathbb{E}}_{x \sim \mathcal{D}}\left[L\left(G_{(f',w_{(G,f')})}(x)\right)\right] + \epsilon_1$) for some $r = r(\epsilon_1)$.*

*Proof.* Theorem 7 proven below guarantees that for any fixed $f \in \mathcal{F}$, with probability $1 - \delta$ that $\mathop{\mathbb{E}}_{x \sim \mathcal{D}}\left[L\left(g_{(f,w_{(G,f)})}(x)\right)\right] < \mathop{\mathbb{E}}_{x \sim \mathcal{D}}\left[L\left(G_{(f,w_{(G,f)})}(x)\right)\right] + \epsilon_1/2$. This can be achieved with $d =$ $(64B^2C_L^2/\epsilon_1^2)\ln(8B/\delta\epsilon_1)$. Now assuming that the above holds, using the above results we can get the following for any $f' \in \mathcal{B}_\infty(f,r) \cap \mathcal{F}$.

$$\begin{aligned}
\mathop{\mathbb{E}}_{x\sim\mathcal{D}}\left[L\left(g_{(f',w_{(g,f')})}(x)\right)\right] &\leq \mathop{\mathbb{E}}_{x\sim\mathcal{D}}\left[L\left(g_{(f,w_{(g,f)})}(x)\right)\right] + C_L rB \\
&\qquad \text{(using Lemma 4.4)} \\
&\leq \mathop{\mathbb{E}}_{x\sim\mathcal{D}}\left[L\left(g_{(f,w_{(G,f)})}(x)\right)\right] + C_L rB \\
&\qquad \text{(using Lemma 4.2)} \\
&\leq \mathop{\mathbb{E}}_{x\sim\mathcal{D}}\left[L\left(G_{(f,w_{(G,f)})}(x)\right)\right] + \epsilon_1/2 + C_L rB \\
&\qquad \text{(using Theorem 7)} \\
&\leq \mathop{\mathbb{E}}_{x\sim\mathcal{D}}\left[L\left(G_{(f',w_{(G,f')})}(x)\right)\right] + \epsilon_1/2 + 2C_L rB \\
&\qquad \text{(using Lemma 4.3)}
\end{aligned}$$

Setting $r = \frac{\epsilon_1}{4C_L B}$ gives us the desired result. $\square$

*Proof.* (of Theorem 3) As mentioned earlier we shall prove the theorem in two parts as follows :

1. (Part I) In this part we shall prove the following :

$$\sup_{f\in\mathcal{F}}\left[\mathop{\mathbb{E}}_{x\sim\mathcal{D}}\left[L\left(g_{(f,w_{(g,f)})}(x)\right)\right] - \mathop{\mathbb{E}}_{x\sim\mathcal{D}}\left[L\left(G_{(f,w_{(G,f)})}(x)\right)\right]\right] \leq \epsilon_1$$

We first set up an $\epsilon$-net over $\mathcal{F}$ at scale $r = \frac{\epsilon_1}{4C_L B}$. Let there be $\mathcal{N}(\mathcal{F},r)$ elements in this net. Taking $d = (64B^2C_L^2/\epsilon_1^2)\ln(8B \cdot \mathcal{N}(\mathcal{F},r)/\delta\epsilon_1)$ landmarks should ensure that the landmarks, with very high probability, are good for all functions in the net by an application of union bound. Since every function in $\mathcal{F}$ is at least $r$-close to some function in the net, Lemma 5 tells us that the same set of landmarks are, with very high probability, good for all the functions in $\mathcal{F}$. This proves the first part of our result.

2. (Part II) In this part we shall prove the following :

$$\sup_{f\in\mathcal{F}}\left[\mathop{\mathbb{E}}_{x\sim\mathcal{D}}\left[L\left(G_{(f,w_{(G,f)})}(x)\right)\right] - \mathop{\mathbb{E}}_{x\sim\mathcal{D}}\left[L\left(g_{(f,w_{(g,f)})}(x)\right)\right]\right] \leq \epsilon_1$$

This part is actually fairly simple to prove. Intuitively, since one can imagine $G$ as being the output of an algorithm that is allowed to take the entire domain as its landmark set, we should expect $\mathop{\mathbb{E}}_{x\sim\mathcal{D}}\left[L\left(G_{(f,w_{(G,f)})}(x)\right)\right] \leq \mathop{\mathbb{E}}_{x\sim\mathcal{D}}\left[L\left(g_{(f,w_{(g,f)})}(x)\right)\right]$ to hold unconditionally for every $f$. For a formal argument, let us build up some more notation. As we have said before, for any transfer function $f$ and arbitrary choice of $d$ landmark pairs $\mathcal{P}$, we let $w_{(g,f)} \in [-B,B]^d$ be the best weighing function for this choice of transfer function and landmark pairs. Now let $\overline{w_{(g,f)}}$ be the best possible extension of $w_{(g,f)}$ to the entire domain. More formally, for any $w^* \in [-B,B]^d$ let $\overline{w^*} = \arg\min_{w\in\mathcal{W}, w|_{\mathcal{P}}=w^*} \mathop{\mathbb{E}}_{x\sim\mathcal{D}}\left[L\left(G_{(f,w)}(x)\right)\right]$.

Now Lemma 4.1 tells us that for any $f \in \mathcal{F}$ and any choice of landmark pairs $\mathcal{P}$, $\mathop{\mathbb{E}}_{x\sim\mathcal{D}}\left[L\left(G_{(f,w_{(G,f)})}(x)\right)\right] \leq \mathop{\mathbb{E}}_{x\sim\mathcal{D}}\left[L\left(G_{(f,\overline{w_{(g,f)}})}(x)\right)\right]$. Furthermore, since $\overline{w_{(g,f)}}$ is chosen to be the most beneficial extension of $w_{(g,f)}$, we also have $\mathop{\mathbb{E}}_{x\sim\mathcal{D}}\left[L\left(G_{(f,\overline{w_{(g,f)}})}(x)\right)\right] \leq \mathop{\mathbb{E}}_{x\sim\mathcal{D}}\left[L\left(g_{(f,w_{(g,f)})}(x)\right)\right]$. Together, these two inequalities give us the second part of the proof. $\square$

This result tells us that in a large enough landmarked space, we shall, for each function $f \in \mathcal{F}$, recover close to the best classifier possible for that transfer function. Thus, if we iterate over the set of transfer functions (or use some gradient-descent based optimization routine), we are bound to select a transfer function that is capable of giving a classifier that is close to the best.

## 2.2 Working with surrogate loss functions

The formulation of a good similarity function suggests a simple learning algorithm that involves the construction of an embedding of the domain into a landmarked space on which the existence of a large margin classifier having low misclassification rate is guaranteed. However, in order to exploit this guarantee we would have to learn the weights $w\left(x_i^+, x_i^-\right)$ associated with this classifier by minimizing the empirical misclassification rate on some training set.

Unfortunately, not only is this problem intractable but also hard to solve approximately [15, 16]. Thus what we require is for the landmarked space to admit a classifier that has low error with respect to a loss function that can also be efficiently minimized on any training set. In such a situation, minimizing the loss on a random training set would, with very high probability, give us weights that give similar performance guarantees as the ones used in the goodness criterion.

With a similar objective in mind, [1] offers variants of its goodness criterion tailored to the hinge loss function which can be efficiently optimized on large training sets (for example LIBSVM [17]). Here we give a general notion of goodness that can be tailored to any arbitrary Lipschitz loss function.

**Definition 6.** *A similarity function $K : \mathcal{X} \times \mathcal{X} \to \mathbb{R}$ is said to be an $(\epsilon, B)$-good similarity for a learning problem with respect to a loss function $L : \mathbb{R} \to \mathbb{R}^+$ where $\epsilon > 0$ if for some transfer function $f : \mathbb{R} \to \mathbb{R}$ and some weighing function $w : \mathcal{X} \times \mathcal{X} \to [-B, B]$, $\mathop{\mathbb{E}}\limits_{x \sim \mathcal{D}} [L(G(x))] \leq \epsilon$ where*
$$G(x) = \mathop{\mathbb{E}}\limits_{x', x'' \sim \mathcal{D} \times \mathcal{D}} \left[ w\left(x', x''\right) f \left( K(x, x') - K(x, x'') \right) | \ell(x') = \ell(x), \ell(x'') \neq \ell(x) \right]$$

One can see that taking the loss functions as $L(x) = \mathbf{1}_{x < C_f \gamma}$ gives us Equation 3 which defines a good similarity under the $0-1$ loss function. It turns out that we can, for any Lipschitz loss function, give similar guarantees on the performance of the classifier in the landmarked space.

**Theorem 7.** *If $K$ is an $(\epsilon, B)$-good similarity function with respect to a $C_L$-Lipschitz loss function $L$ then for any $\epsilon_1 > 0$, with probability at least $1 - \delta$ over the choice of $d = (16B^2 C_L^2 / \epsilon_1^2) \ln(4B/\delta \epsilon_1)$ positive and negative samples from $\mathcal{D}^+$ and $\mathcal{D}^-$ respectively, the expected loss of the classifier $g(x)$ with respect to $L$ satisfies $\mathop{\mathbb{E}}\limits_{x} [L(g(x))] \leq \epsilon + \epsilon_1$ where $g(x) = \frac{1}{d} \sum_{i=1}^{d} w\left(x_i^+, x_i^-\right) f\left(K(x, x_i^+) - K(x, x_i^-)\right)$.*

*Proof.* For any $x \in \mathcal{X}$, we have, by an application of Hoeffding bounds $\mathop{\Pr}\limits_{g} [|G(x) - g(x)| > \epsilon_1] < 2 \exp\left(-\frac{\epsilon_1^2 d}{2B^2}\right)$ since $|g(x)| \leq B$. Here the notation $\mathop{\Pr}\limits_{g}$ signifies that the probability is over the choice of the landmark points. Thus for $d > \frac{4B^2}{\epsilon_1^2} \ln\left(\frac{2}{\delta}\right)$, we have $\mathop{\Pr}\limits_{g} [|G(x) - g(x)| > \epsilon_1] < \delta^2$. For sake of simplicity let us denote by $\mathsf{BAD}(x)$ the event $|G(x) - g(x)| > \epsilon_1$. Thus we have, for every $x \in \mathcal{X}$, $\mathop{\mathbb{E}}\limits_{g} \left[\mathbf{1}_{\mathsf{BAD}(x)}\right] < \delta^2$. Since this is true for every $x \in \mathcal{X}$, this also holds in expectation i.e. $\mathop{\mathbb{E}}\limits_{x} \mathop{\mathbb{E}}\limits_{g} \left[\mathbf{1}_{\mathsf{BAD}(x)}\right] < \delta^2$. The expectation over $x$ is with respect to the problem distribution $\mathcal{D}$. Applying Fubini's Theorem gives us $\mathop{\mathbb{E}}\limits_{g} \mathop{\mathbb{E}}\limits_{x} \left[\mathbf{1}_{\mathsf{BAD}(x)}\right] < \delta^2$ which upon application of Markov's inequality gives us $\mathop{\Pr}\limits_{g} \left[\mathop{\mathbb{E}}\limits_{x} \left[\mathbf{1}_{\mathsf{BAD}(x)}\right] > \delta\right] < \delta$. Thus, with very high probability we would always choose landmarks such that $\mathop{\Pr}\limits_{x} [BAD(x)] < \delta$. Thus we have, in such a situation, $\mathop{\mathbb{E}}\limits_{x} [|G(x) - g(x)|] \leq (1-\delta)\epsilon_1 + \delta \cdot 2B$ since $\sup\limits_{x \in \mathcal{X}} |G(x) - g(x)| \leq 2B$. For small enough $\delta$ we have $\mathop{\mathbb{E}}\limits_{x} [|G(x) - g(x)|] \leq 2\epsilon_1$.

Thus we have $\mathop{\mathbb{E}}\limits_{x} [L(g(x))] - \mathop{\mathbb{E}}\limits_{x} [L(G(x))] = \mathop{\mathbb{E}}\limits_{x} [L(g(x)) - L(G(x))] \leq \mathop{\mathbb{E}}\limits_{x} [C_L \cdot |g(x) - G(x)|] = C_L \cdot \mathop{\mathbb{E}}\limits_{x} [|g(x) - G(x)|] \leq 2C_L \epsilon_1$ where we used the Lipschitz properties of the loss function $L$ to arrive at the second inequality. Putting $\epsilon_1 = \frac{\epsilon'_1}{2C_L}$ we have $\mathop{\mathbb{E}}\limits_{x} [L(g(x))] \leq \mathop{\mathbb{E}}\limits_{x} [L(G(x))] + \epsilon'_1 \leq \epsilon + \epsilon'_1$ which gives us our desired result.

**Algorithm 1** DSELECT

**Require:** A training set $T$, landmarking size $d$.
**Ensure:** A set of $d$ landmark pairs/singletons.
1: $Ł \leftarrow \text{get-random-element}(T)$, $\mathcal{P}_{\text{FTUNE}} \leftarrow \emptyset$
2: **for** $j = 2$ **to** $d$ **do**
3: $\quad z \leftarrow \arg\min_{x \in T} \sum_{x' \in Ł} K(x, x')$.
4: $\quad Ł \leftarrow Ł \cup \{z\}$, $T \leftarrow T \setminus \{z\}$
5: **end for**
6: **for** $j = 1$ **to** $d$ **do**
7: $\quad$ Sample $z_1, z_2$, s.t., $\ell(z_1) = 1$, $\ell(z_2) = -1$ randomly from $Ł$ with replacement
8: $\quad \mathcal{P}_{\text{FTUNE}} \leftarrow \mathcal{P}_{\text{FTUNE}} \cup \{(z_1, z_2)\}$
9: **end for**
10: **return** $Ł$ (for BBS), $\mathcal{P}_{\text{FTUNE}}$ (for FTUNE)

**Algorithm 2** FTUNE

**Require:** A family of transfer functions $\mathcal{F}$, a similarity function $K$ and a loss function $L$
**Ensure:** An optimal transfer function $f^* \in \mathcal{F}$.
1: Select $d$ landmark pairs $\mathcal{P}$.
2: **for all** $f \in \mathcal{F}$ **do**
3: $\quad w_f \leftarrow \text{TRAIN}(\mathcal{P}, L)$, $L_f \leftarrow L(w_f)$
4: **end for**
5: $f^* \leftarrow \arg\min_{f \in \mathcal{F}} L_f$
6: **return** $(f^*, w_{f^*})$.

Actually we can prove something stronger since $\left|\mathbb{E}_x[L(g(x))] - \mathbb{E}_x[L(G(x))]\right| = \left|\mathbb{E}_x[L(g(x)) - L(G(x))]\right| \leq \mathbb{E}_x[|L(g(x)) - L(G(x))|] \leq \mathbb{E}_x[C_L \cdot |g(x) - G(x)|] \leq \epsilon'_1$. Thus we have $\epsilon - \epsilon'_1 \leq \mathbb{E}_x[L(g(x))] \leq \epsilon + \epsilon'_1$. $\square$

If the loss function is hinge loss at margin $\gamma$ then $C_L = \frac{1}{\gamma}$. The $0 - 1$ loss function and the loss function $L(x) = \mathbb{1}_{x < \gamma}$ (implicitly used in Definition 1 and Theorem 2) are not Lipschitz and hence this proof technique does not apply to them.

### 2.3 Selecting informative landmarks

Recall that the generalization guarantees we described in the previous section rely on random selection of landmark pairs from a fixed distribution over the domain. However, in practice, a totally random selection might require one to select a large number of landmarks, thereby leading to an inefficient classifier in terms of training as well as test times. For typical domains such as computer vision, similarity function computation is an expensive task and hence selection of a small number of landmarks should lead to a significant improvement in the test times. For this reason, we propose a landmark pair selection heuristic which we call DSELECT (see Algorithm 1). The heuristic generalizes naturally to multi-class problems and can also be applied to the classification model of Balcan-Blum that uses landmark singletons instead of pairs.

At the core of our heuristic is a novel notion of *diversity* among landmarks. Assuming $K$ is a normalized similarity kernel, we call a set of points $S \subset \mathcal{X}$ *diverse* if the average inter-point similarity is small i.e $\frac{1}{|S|(|S|-1)} \sum_{x,y \in S, x \neq y} K(x, y) \ll 1$ (in case we are working with a distance kernel we would require large inter-point distances). The key observation behind DSELECT is that a non-diverse set of landmarks would cause all data points to receive identical embeddings and linear separation would be impossible. Small inter-landmark similarity, on the other hand would imply that the landmarks are well-spread in the domain and can capture novel patterns in the data.

Similar notions of diversity have been used in the past for ensemble classifiers [18] and $k$-NN classifiers [5]. Here we use this notion to achieve a better embedding into the landmarked space. Experimental results demonstrate that the heuristic offers significant performance improvements over random landmark selection (see Figure 1). One can easily extend Although Algorithm 1 to multi-class problems by selecting a fixed number of landmarks from each class.

## 3 Empirical results

In this section, we empirically study the performance of our proposed methods on a variety of benchmark datasets. We refer to the algorithmic formulation presented in [1] as BBS and its augmentation using DSELECT as BBS+D. We refer to the formulation presented in [2] as DBOOST. We refer to our transfer function learning based formulation as FTUNE and its augmentation using DSELECT as FTUNE+D. In multi-class classification scenarios we will use a *one-vs-all* formulation which

| Dataset/Method | BBS | DBOOST | FTUNE+D-S | Dataset/Method | BBS | DBOOST | FTUNE+D-S |
|---|---|---|---|---|---|---|---|
| AmazonBinary | 0.73(0.13) | 0.77(0.10) | **0.84**(0.12) | AmazonBinary | 0.78(0.11) | 0.82(0.10) | **0.88**(0.07) |
| AuralSonar | **0.82**(0.08) | **0.81**(0.08) | **0.80**(0.08) | AuralSonar | **0.88**(0.06) | 0.85(0.07) | 0.85(0.07) |
| Patrol | 0.51(0.06) | 0.34(0.11) | **0.58**(0.06) | Patrol | **0.79**(0.05) | 0.55(0.12) | **0.79**(0.07) |
| Voting | **0.95**(0.03) | **0.94**(0.03) | **0.94**(0.04) | Voting | **0.97**(0.02) | **0.97**(0.01) | **0.97**(0.02) |
| Protein | 0.98(0.02) | **1.00**(0.01) | 0.98(0.02) | Protein | **0.98**(0.02) | **0.99**(0.02) | **0.98**(0.02) |
| Mirex07 | 0.12(0.01) | 0.21(0.03) | **0.28**(0.03) | Mirex07 | 0.17(0.02) | 0.31(0.04) | **0.35**(0.02) |
| Amazon47 | 0.39(0.06) | 0.07(0.04) | **0.61**(0.08) | Amazon47 | 0.40(0.13) | 0.07(0.05) | **0.66**(0.07) |
| FaceRec | 0.20(0.04) | 0.12(0.03) | **0.63**(0.04) | FaceRec | 0.27(0.05) | 0.19(0.03) | **0.64**(0.04) |

(a) 30 Landmarks    (b) 300 Landmarks

Table 1: Accuracies for Benchmark Similarity Learning Datasets for Embedding Dimensionality=30, 300. Bold numbers indicate the best performance with $95\%$ confidence level.

presents us with an opportunity to further exploit the transfer function by learning separate transfer function per class (i.e. per one-vs-all problem). We shall refer to our formulation using a single (resp. multiple) transfer function as FTUNE+D-S (resp. FTUNE+D-M). We take the class of *ramp* functions indexed by a slope parameter as our set of transfer functions. We use 6 different values of the slope parameter $\{1, 5, 10, 50, 100, 1000\}$. Note that these functions (approximately) include both the identity function (used by [1]) and the sign function (used by [2]).

Our goal in this section is two-fold: 1) to show that our FTUNE method is able to learn a more suitable transfer function for the underlying data than the existing methods BBS and DBOOST and 2) to show that our diversity based heuristic for landmark selection performs better than random selection. To this end, we perform experiments on a few benchmark datasets for learning with similarity (non-PSD) functions [5] as well as on a variety of standard UCI datasets where the similarity function used is the Gaussian kernel function.

For our experiments, we implemented our methods FTUNE and FTUNE+D as well as BBS and BBS+D using MATLAB while using LIBLINEAR [10] for SVM classification. For DBOOST, we use the C++ code provided by the authors of [2]. On all the datasets we randomly selected a fixed percentage of data for training, validation and testing. Except for DBOOST, we selected the SVM penalty constant $C$ from the set $\{1, 10, 100, 1000\}$ using validation. For each method and dataset, we report classification accuracies averaged over 20 runs. We compare accuracies obtained by different methods using $t$-test at $95\%$ significance level.

### 3.1 Similarity learning datasets

First, we conduct experiments on a few similarity learning datasets [5]; these datasets provide a (non-PSD) similarity matrix along with class labels. For each of the datasets, we randomly select $70\%$ of the data for training, $10\%$ for validation and the remaining for testing purposes. We then apply our FTUNE-S, FTUNE+D-S, BBS+D methods along with BBS and DBOOST with varying number of landmark pairs. Note that we do not apply our FTUNE-M method to these datasets as it overfits heavily to these datasets as typically they are small in size.

We first compare the accuracy achieved by FTUNE+D-S with the existing methods. Table 1 compares the accuracies achieved by our FTUNE+D-S method with those of BBS and DBOOST over different datasets when using landmark sets of sizes 30 and 300. Numbers in brackets denote standard deviation over different runs. Note that in both the tables FTUNE+D-S is one of the best methods (upto $95\%$ significance level) on all but one dataset. Furthermore, for datasets with large number of classes such as Amazon47 and FaceRec our method outperforms BBS and DBOOST by at least $20\%$ percent. Also, note that some of the datasets have multiple bold faced methods, which means that the two sample $t$-test (at $95\%$ level) rejects the hypothesis that their mean is different.

Next, we evaluate the effectiveness of our landmark selection criteria for both BBS and our method. Figure 1 shows the accuracies achieved by various methods on four different datasets with increasing number of landmarks. Note that in all the datasets, our diversity based landmark selection criteria increases the classification accuracy by around $5 - 6\%$ for small number of landmarks.

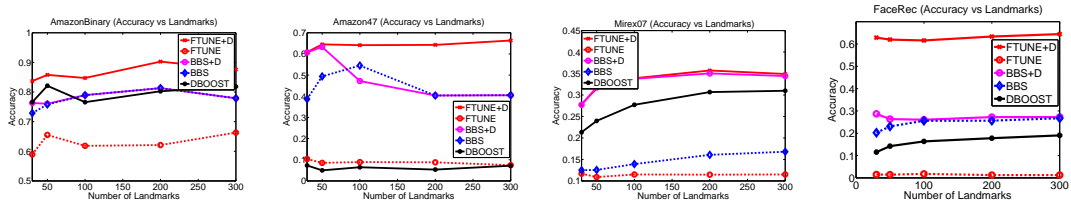

Figure 1: Accuracy obtained by various methods on four different datasets as the number of landmarks used increases. Note that for small number of landmarks (30, 50) our diversity based landmark selection criteria increases accuracy for both BBS and our method FTUNE-S significantly.

| Dataset/Method | BBS | DBOOST | FTUNE-S | FTUNE-M | Dataset/Method | BBS | DBOOST | FTUNE-S | FTUNE-M |
|---|---|---|---|---|---|---|---|---|---|
| Cod-rna | 0.93(0.01) | 0.89(0.01) | **0.93**(0.01) | **0.93**(0.01) | Cod-rna | **0.94**(0.00) | 0.93(0.00) | **0.94**(0.00) | **0.94**(0.00) |
| Isolet | 0.81(0.01) | 0.67(0.01) | **0.84**(0.01) | 0.83(0.01) | Isolet | 0.91(0.01) | 0.89(0.01) | **0.93**(0.01) | **0.93**(0.00) |
| Letters | 0.67(0.02) | 0.58(0.01) | **0.69**(0.01) | 0.68(0.02) | Letters | 0.72(0.01) | **0.84**(0.01) | 0.83(0.01) | 0.83(0.01) |
| Magic | 0.82(0.01) | 0.81(0.01) | **0.84**(0.01) | 0.84(0.01) | Magic | 0.84(0.01) | 0.84(0.00) | **0.85**(0.01) | **0.85**(0.01) |
| Pen-digits | 0.94(0.01) | 0.93(0.01) | **0.97**(0.01) | **0.97**(0.01) | Pen-digits | 0.96(0.00) | **0.99**(0.00) | **0.99**(0.00) | **0.99**(0.00) |
| Nursery | **0.91**(0.01) | **0.91**(0.01) | 0.90(0.01) | 0.90(0.00) | Nursery | 0.93(0.01) | **0.97**(0.00) | 0.96(0.00) | **0.97**(0.00) |
| Faults | **0.70**(0.01) | 0.68(0.02) | **0.70**(0.02) | **0.71**(0.02) | Faults | 0.72(0.02) | **0.74**(0.02) | 0.73(0.02) | 0.73(0.02) |
| Mfeat-pixel | **0.94**(0.01) | 0.91(0.01) | **0.95**(0.01) | 0.94(0.01) | Mfeat-pixel | 0.96(0.01) | **0.97**(0.01) | **0.97**(0.01) | **0.97**(0.01) |
| Mfeat-zernike | **0.79**(0.02) | 0.72(0.02) | **0.79**(0.02) | **0.79**(0.02) | Mfeat-zernike | 0.81(0.01) | 0.79(0.01) | **0.82**(0.02) | **0.82**(0.01) |
| Opt-digits | 0.92(0.01) | 0.89(0.01) | **0.94**(0.01) | **0.94**(0.01) | Opt-digits | 0.95(0.01) | 0.97(0.00) | **0.98**(0.01) | **0.98**(0.00) |
| Satellite | 0.85(0.01) | 0.86(0.01) | 0.86(0.01) | **0.87**(0.01) | Satellite | 0.85(0.01) | **0.90**(0.01) | 0.89(0.01) | 0.89(0.01) |
| Segment | 0.90(0.01) | **0.93**(0.01) | 0.92(0.01) | 0.92(0.01) | Segment | 0.90(0.01) | **0.96**(0.01) | 0.96(0.01) | 0.96(0.01) |

(a) 30 Landmarks      (b) 300 Landmarks

Table 2: Accuracies for Gaussian Kernel for Embedding Dimensionality=30. Bold numbers indicate the best performance with 95% confidence level. Note that both our methods, especially FTUNE-S, performs significantly better than the existing methods.

### 3.2 UCI benchmark datasets

We now compare our FTUNE method against existing methods on a variety of UCI datasets [19]. We ran experiments with FTUNE and FTUNE+D but the latter did not provide any advantage. So for lack of space we drop it from our presentation and only show results for FTUNE-S (FTUNE with a single transfer function) and FTUNE-M (FTUNE with one transfer function per class). Similar to [2], we use the Gaussian kernel function as the similarity function for evaluating our method. We set the "width" parameter in the Gaussian kernel to be the mean of all pair-wise training data distances, a standard heuristic. For all the datasets, we randomly select 50% data for training, 20% for validation and the remaining for testing. We report accuracy values averaged over 20 runs for each method with varying number of landmark pairs.

Table 2 compares the accuracies obtained by our FTUNE-S and FTUNE-M methods with those of BBS and DBOOST when applied to different UCI benchmark datasets. Note that FTUNE-S is one of the best on most of the datasets for both the landmarking sizes. Also, BBS performs reasonably well for small landmarking sizes while DBOOST performs well for large landmarking sizes. In contrast, our method consistently outperforms the existing methods in both the scenarios.

Next, we study accuracies obtained by our method for different landmarking sizes. Figure 2 shows accuracies obtained by various methods as the number of landmarks selected increases. Note that the accuracy curve of our method dominates the accuracy curves of all the other methods, i.e. our method is consistently better than the existing methods for all the landmarking sizes considered.

### 3.3 Discussion

We note that since FTUNE selects its output by way of validation, it is susceptible to over-fitting on small datasets but at the same time, capable of giving performance boosts on large ones. We observe a similar trend in our experiments – on smaller datasets (such as those in Table 1 with average dataset size 660), FTUNE over-fits and performs worse than BBS and DBOOST. However, even in these cases, DSELECT (intuitively) removes redundancies in the landmark points thus allowing FTUNE to recover the best transfer function. In contrast, for larger datasets like those in Table 2 (average size 13200), FTUNE is itself able to recover better transfer functions than the baseline methods and hence both FTUNE-S and FTUNE-M perform significantly better than the baselines. Note that DSELECT is not able to provide any advantage here since the datasets sizes being large, greedy selection actually ends up hurting the accuracy.

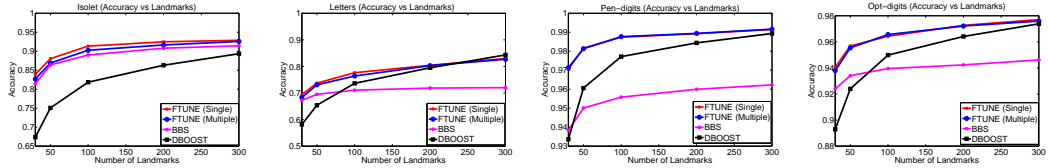

Figure 2: Accuracy achieved by various methods on four different UCI repository datasets as the number of landmarks used increases. Note that both FTUNE-S and FTUNE-M perform significantly better than BBS and DBOOST for small number of landmarks (30, 50).

## Acknowledgments

We thank the authors of [2] for providing us with C++ code of their implementation. P. K. is supported by Microsoft Corporation and Microsoft Research India under a Microsoft Research India Ph.D. fellowship award. Most of this work was done while P. K. was visiting Microsoft Research Labs India, Bangalore.

## A  Comparison with the models of Balcan-Blum and Wang *et al*

In [2], Wang *et al* consider a model of learning with distance functions. Their model is similar to our but for the difference that they restrict themselves to the use of a single transfer function namely the sign function $f = \text{sgn}()$. More formally they have the following notion of a *good distance function*.

**Definition 8** ([1] Definition 4). *A distance function $\mathcal{X}$, $d : \mathcal{X} \times \mathcal{X} \to \mathbb{R}$ is said to be an $(\epsilon, \gamma, B)$-good distance for a learning problem where $\epsilon, \gamma, B > 0$ if there exist two class conditional probability distributions $\tilde{\mathcal{D}}(x|\ell(x) = 1)$ and $\tilde{\mathcal{D}}(x|\ell(x) = -1)$ such that for all $x \in \mathcal{X}$, $\frac{\tilde{\mathcal{D}}(x|\ell(x)=1)}{\mathcal{D}(x|\ell(x)=1)} < \sqrt{B}$ and $\frac{\tilde{\mathcal{D}}(x|\ell(x)=-1)}{\mathcal{D}(x|\ell(x)=-1)} < \sqrt{B}$ where $\mathcal{D}(x|\ell(x) = 1)$ and $\mathcal{D}(x|\ell(x) = -1)$ are the class conditional probability distributions of the problem, such that at least a $1 - \epsilon$ probability mass of examples $x \sim \mathcal{D}$ satisfies*

$$\underset{x',x'' \sim \tilde{\mathcal{D}} \times \tilde{\mathcal{D}}}{\tilde{\mathcal{D}}} [d(x,x') < d(x,x'') | \ell(x') = \ell(x), \ell(x'') \neq \ell(x)] \geq \frac{1}{2} + \gamma \quad (4)$$

It can be shown (and is implicit in the proof of Theorem 5 in [2]) that the above condition is equivalent to

$$\underset{x',x'' \sim \mathcal{D} \times \mathcal{D}}{\mathbb{E}} \left[ w_{\ell(x)}(x') w_{-\ell(x)}(x'') \text{sgn} \left( d(x,x'') - d(x,x') \right) | \ell(x') = \ell(x), \ell(x'') \neq \ell(x) \right] \geq 2\gamma$$

where $w_1(x) := \frac{\tilde{\mathcal{D}}(x|\ell(x)=1)}{\mathcal{D}(x|\ell(x)=1)}$ and $w_{-1}(x) := \frac{\tilde{\mathcal{D}}(x|\ell(x)=-1)}{\mathcal{D}(x|\ell(x)=-1)}$. Now define $\varpi(x',x'') := w_{\ell(x')}(x') w_{\ell(x'')}(x'')$ and take $f = \text{sgn}()$ as the transfer function in our model. We have, for a $1 - \epsilon$ fraction of points,

$$\underset{x',x'' \sim \mathcal{D} \times \mathcal{D}}{\mathbb{E}} \left[ \varpi \left( x',x'' \right) f \left( K(x,x') - K(x,x'') \right) | \ell(x') = \ell(x), \ell(x'') \neq \ell(x) \right] \geq C_f \gamma$$

which is clearly seen to be equivalent to

$$\underset{x',x'' \sim \mathcal{D} \times \mathcal{D}}{\mathbb{E}} \left[ w_{\ell(x)}(x') w_{-\ell(x)}(x'') \text{sgn} \left( K(x,x') - K(x,x'') \right) | \ell(x') = \ell(x), \ell(x'') \neq \ell(x) \right] \geq \gamma$$

since $C_f = 1$ for the sgn() function. Thus the Wang *et al* model of learning is an instantiation of our proposed model.

In [1], Balcan-Blum present a model of learning with similarity functions. Their model does not consider landmark pairs, just singletons. Accordingly, instead of assigning a weight to each landmark pair, one simply assigns a weight to each element of the domain. Consequently one arrives at the following notion of a *good similarity*.

**Definition 9** ([2], Definition 3). *A similarity measure $K : \mathcal{X} \times \mathcal{X} \to \mathbb{R}$ is said to be an $(\epsilon, \gamma)$-good similarity for a learning problem where $\epsilon, \gamma > 0$ if for some weighing function $w : \mathcal{X} \to [-1, 1]$, at least a $1 - \epsilon$ probability mass of examples $x \sim \mathcal{D}$ satisfies*

$$\underset{x' \sim \mathcal{D}}{\mathbb{E}} [w(x') K(x,x') | \ell(x') = \ell(x)] \geq \underset{x' \sim \mathcal{D}}{\mathbb{E}} [w(x') K(x,x') | \ell(x') \neq \ell(x)] + \gamma \quad (5)$$

Now define $w_+ := \mathop{\mathbb{E}}\limits_{x \sim \mathcal{D}}[w(x)|\ell(x) = 1]$ and $w_- := \mathop{\mathbb{E}}\limits_{x \sim \mathcal{D}}[w(x)|\ell(x) = -1]$. Furthermore, take $\varpi(x', x'') = w(x')w(x'')$ as the weight function and $f = \text{id}()$ as the transfer function in our model. Then we have, for a $1 - \epsilon$ fraction of the points,

$$\mathop{\mathbb{E}}\limits_{x',x'' \sim \mathcal{D} \times \mathcal{D}}[\varpi(x', x'') f(K(x, x') - K(x, x''))|\ell(x') = \ell(x), \ell(x'') \neq \ell(x)] \geq C_f \gamma$$

$$\equiv \mathop{\mathbb{E}}\limits_{x',x'' \sim \mathcal{D} \times \mathcal{D}}[\varpi(x', x'')(K(x, x') - K(x, x''))|\ell(x') = \ell(x), \ell(x'') \neq \ell(x)] \geq \gamma$$

$$\equiv \mathop{\mathbb{E}}\limits_{x',x'' \sim \mathcal{D} \times \mathcal{D}}[\varpi(x', x'') K(x, x')|\ell(x') = \ell(x), \ell(x'') \neq \ell(x)] \geq$$

$$\mathop{\mathbb{E}}\limits_{x',x'' \sim \mathcal{D} \times \mathcal{D}}[\varpi(x', x'') K(x, x'')|\ell(x') = \ell(x), \ell(x'') \neq \ell(x)] + \gamma$$

$$\equiv w_{-\ell(x)} \mathop{\mathbb{E}}\limits_{x' \sim \mathcal{D}}[w(x')K(x, x')|\ell(x') = \ell(x)] \geq w_{\ell(x)} \mathop{\mathbb{E}}\limits_{x' \sim \mathcal{D}}[w(x')K(x, x')|\ell(x') \neq \ell(x)] + \gamma$$

$$\equiv \mathop{\mathbb{E}}\limits_{x' \sim \mathcal{D}}[w'(x')K(x, x')|\ell(x') = \ell(x)] \geq \mathop{\mathbb{E}}\limits_{x' \sim \mathcal{D}}[w'(x')K(x, x')|\ell(x') \neq \ell(x)] + \gamma$$

where $C_f = 1$ for the id() function and $w'(x) = w(x)w_{-\ell(x)}$. Note that this again guarantees a classifier with margin $\gamma$ in the landmarked space. Thus the Balcan-Blum model can also be derived in our model.